\documentclass[10pt,twocolumn,letterpaper]{article}

\usepackage[pagenumbers]{cvpr} 

\usepackage{multirow}
\usepackage{array}
\usepackage{color}
\usepackage{colortbl}
\usepackage{tabularx} 
\usepackage{pifont}
\usepackage{wrapfig}
\usepackage{booktabs} 

\definecolor{mycolor_blue}{HTML}{E7EFFA}
\definecolor{mycolor_green}{HTML}{E6F8E0}
\definecolor{mycolor_gray}{HTML}{ECECEC}
\definecolor{pearDark}{HTML}{2980B9}
\definecolor{cvprblue}{rgb}{0.21,0.49,0.74}
\usepackage[pagebackref,breaklinks,colorlinks,allcolors=cvprblue]{hyperref}

\def\paperID{1579} 
\def\confName{CVPR}
\def\confYear{2025}

\definecolor{scolor}{RGB}{111,168,220}
\definecolor{hcolor}{RGB}{111,176,81}
\definecolor{ocolor}{RGB}{224,103,102}
\definecolor{wcolor}{RGB}{246,178,107}

\title{\textcolor{scolor}{D}\textcolor{hcolor}{o}\textcolor{ocolor}{r}\textcolor{wcolor}{a}Cycle: \textcolor{scolor}{D}omain-\textcolor{hcolor}{O}\textcolor{ocolor}{r}iented \textcolor{wcolor}{A}daptation of Unified Generative Model \\ in Multimodal Cycles}

\author{Rui Zhao, Weijia Mao, Mike Zheng Shou\thanks{Corresponding Author.}\\ 
Show Lab, National University of Singapore 
}

\begin{document}

\maketitle
\begin{abstract}

Adapting generative models to specific domains presents an effective solution for satisfying specialized requirements. However, adapting to some complex domains remains challenging, especially when these domains require substantial paired data to capture the targeted distributions. Since unpaired data from a single modality, such as vision or language, is more readily available, we utilize the bidirectional mappings between vision and language learned by the unified generative model to enable training on unpaired data for domain adaptation. Specifically, we propose DoraCycle, which integrates two multimodal cycles: text-to-image-to-text and image-to-text-to-image. The model is optimized through cross-entropy loss computed at the cycle endpoints, where both endpoints share the same modality. This facilitates self-evolution of the model without reliance on annotated text-image pairs. Experimental results demonstrate that for tasks independent of paired knowledge, such as stylization, DoraCycle can effectively adapt the unified model using only unpaired data. For tasks involving new paired knowledge, such as specific identities, a combination of a small set of paired image-text examples and larger-scale unpaired data is sufficient for effective domain-oriented adaptation. 
The code will be released at \href{https://github.com/showlab/DoraCycle}{https://github.com/showlab/DoraCycle}.

\end{abstract}    
\section{Introduction}

The adaptation of pre-trained generative models to specific domains is an important aspect of advancing personalized content creation, from stylized media outputs to customized identity generation~\cite {ruiz2022dreambooth, wang2024instantid}. 
However, effectively adapting generative models to complex domains remains challenging, particularly when these domains require extensive amounts of paired data to accurately capture the desired distributions.
For instance, learning the visual styles and character identities across a unique movie, which involves understanding multiple characters, their relationships, and diverse settings, is a highly complex task that demands vast amounts of paired frames and captions data. 
Collecting such paired data, especially for multimodal tasks involving vision and language, is often laborious and impractical, limiting the potential to adapt generative models at scale.

\begin{figure}[t]  
    \centering
\includegraphics[width=1.0\linewidth]{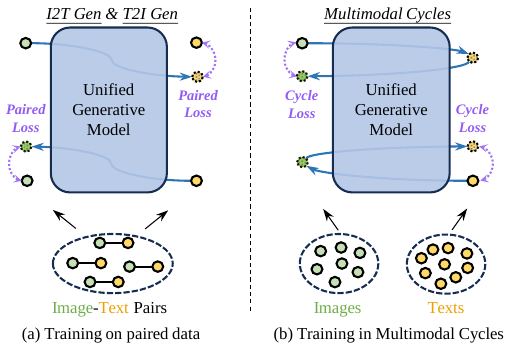}
    \caption{Training paradigms for unified generative models. (a) Traditional training involves using paired image-text data and optimizing the unified model with paired losses for both image-to-text (I2T) and text-to-image (T2I) generation tasks. (b) In contrast, the proposed multimodal cycle training framework leverages unpaired images and texts. By using cycle-consistency losses, the unified model learns to maintain consistency between input and output across modalities, enabling adaptation without the need for extensive paired datasets.}
\label{fig:pipeline_overview}
\end{figure}

High-quality image-text paired data is relatively rare and scarce, but unpaired images and texts are readily available in our daily lives, such as video websites, image platforms, and content from novel websites. Therefore, we aim to explore \textit{whether it is possible to adapt generative models to target domains based on unpaired data}. To achieve this goal, it is crucial for the model to have an internal capability to align the two modalities to some extent. Fortunately, recent advancements in unified generative models have encouraged us to pursue this direction.

Recent advanced unified generative models~\citep{seed-x, dreamllm, team2024chameleon, zhou2024transfusion, xie2024show} have shown great potential in unifying multimodal understanding and generation within a single model. These unified models are capable of processing and generating content across different modalities, \textit{i.e.} vision and language, within a shared framework. 
By leveraging the bidirectional mappings between vision and language, which are learned by the unified generative model in its pre-training stage, we can map each data from one modality to another and then back to the original modality, as shown in Fig.~\ref{fig:pipeline_overview}. Through these two mappings, data can be maintained within the same modality, thereby imposing constraints on the deviation introduced in the process. This only requires computing the cross-entropy loss between the data in the same modality, without any paired data supervision.

To this end, we introduce DoraCycle, a framework for adapting unified multimodal generative models to target domains, through cycle-consistent learning with unpaired data. Unlike previous adaptation methods that heavily rely on paired text-image data, the proposed framework leverages the shared latent space of unified models to learn consistent transformations between modalities without requiring paired training examples. Specifically, we design two cycles, \textit{i.e.} text-to-image-to-text (T cycle) and image-to-text-to-image (I cycle).
As shown by Fig.~\ref{fig:pipeline_overview} (b), leveraging the pre-trained vision-language alignment of the unified model, each multimodal cycle involves two cross-modality mappings while optimizing in the same modality. This enables calculating loss on unpaired data while implicitly refining cross-modal alignment through the intermediate step.

In practice, since there is no labeled ground-truth, mapping data to another modality requires multi-step inference, such as predicting the next token multiple times for text generation. Allowing all forward steps to participate in gradient backpropagation can lead to a catastrophic gradient explosion. Therefore, we stop gradients during multi-step inference and use the generated data as pseudo labels for the model to forward once again, allowing gradients to propagate.
Moreover, we found that since a complete cycle requires the model to forward twice, it can lead to training instability, with the quality of pseudo data generated in the middle being compromised. To enhance the stability of pseudo data generation, we maintain a slowly updated EMA (Exponential Moving Average)~\citep{tarvainen2017mean} model, which is used for inference to generate pseudo data. Additionally, we employ gradient clipping techniques to avoid conflicts in the optimization directions of the two cycles, further increasing training stability.

The experiments indicate that for tasks independent of paired knowledge, such as stylization and domain-specific adaptation, DoraCycle can adapt the unified model with only unpaired data, which is both more practical and scalable. For tasks that require new paired knowledge, such as identity-specific adaptation, DoraCycle effectively utilizes small amounts of paired data along with larger-scale unpaired data, making it a flexible solution for generative adaptation challenges.
We conduct extensive experiments that compare DoraCycle to existing methods, showing that our approach achieves comparable or superior results while significantly reducing the need for paired data. 
This ability to harness large-scale unpaired data, combined with strategic usage of small paired datasets, makes DoraCycle a feasible solution for personalized multimodal content generation across a wide range of applications.
\section{Related Works}

\subsection{Multimodal Generation and Understanding}

Generating visual contents from text and describing them through natural language have been extensively studied as core multimodal tasks. Advanced generative models~\citep{chen2020generative, rombach2022high,ramesh2022hierarchical,dalle2,pixart,dai2023emu,saharia2022photorealistic,esser2024scaling,balaji2022ediffi, zhao2023zero, gu2023mix, chen2024pixart-sigma, ma2024sit, zhao2024motiondirector,ma2024latte, dhariwal2021diffusion, zhao2025evolvedirector}, such as DALL·E~\citep{dalle, ramesh2022dalle2}, Stable Diffusion~\citep{rombach2022highresolution}, demonstrate remarkable generation capabilities, producing high-quality and diverse contents from textual prompts. Meanwhile, image captioning models~\cite{hossain2019comprehensive, wang2022git, wang2022ofa, hu2022scaling, hu2023exploiting, wang2024ladic}, such as mPLUG~\citep{li2022mplug}, and BLIP~\citep{li2023blip}, push the boundaries of visual understanding, generating accurate and context-aware descriptions. Additionally, recent advancements in multimodal large language models~\cite{li2024multimodal}, such as LLaVA~\citep{llava}, MiniGPT-4~\citep{minigpt4}, and InstructBLIP~\citep{instructblip},  have significantly improved the ability to understand and reason about visual content.

Besides the powerful foundational generative models, adapting or customizing them attracts increasing interest, which enables more personalized and specific outputs based on user preferences~\cite{gal2022image, mokady2023null, jia2023taming, chen2023photoverse, chen2024anydoor, kumari2023multi, guo2023zero, yang2023one}. Approaches like DreamBooth~\cite{ruiz2022dreambooth} enable user-specific customization by fine-tuning generative models with personal data, allowing the generation of content tailored to individual needs or preferences.

\subsection{Unified Multimodal Generative Models}

Unified multimodal generative models aim to bridge the gap between understanding and generation tasks, and integrate vision and language into a single framework, enabling the model to learn shared representations across modalities~\citep{aghajanyan2022cm3, yu2023scaling, you2023cobit, seed-x,wu2023next, CoDI, x-vila,dreamllm,jointly, xie2024show, zhou2024transfusion,  wu2024janus, sun2024generative}.  
SEED-X~\citep{seed-x} utilizes a unified architecture where visual features extracted from the CLIP ViT encoder~\citep{radford2021learning} are combined with text tokens and fed into a large language model to enable both next-word prediction and image regression tasks.
DreamLLM~\citep{dreamllm} extends the generative capability of large language models by combining multimodal inputs directly into LLMs.
Chameleon~\citep{team2024chameleon} employs a discrete tokenization approach for both visual and textual inputs, converting all modalities into a unified token space that is processed by a transformer-based architecture. 
Transfusion~\citep{zhou2024transfusion} introduces an advanced integration mechanism that focuses on directly fusing visual encoding with language tokens, allowing the model to effectively translate visual information into textual formats while maintaining the semantic integrity of both modalities.
Show-o~\citep{xie2024show} combines autoregressive modeling with a discrete diffusion process, enabling the generation of high-quality outputs that are aligned across modalities. Our work leverages the advancements made by these foundational models and explores how to adapt the foundational model to specific domains.

\subsection{Cycle Consistency}
Cycle consistency has been utilized in computer vision and natural language processing as a means to enhance model robustness and consistency~\cite{he2016dual, hoffman2018cycada, shah2019cyclevqa, choi2018stargan, dwibedi2019temporal, zhang2021few, shah2019cycle, shah2019cycle}. CycleGAN~\citep{zhu2017cyclegan}  introduced cycle consistency loss to align unpaired image domains, ensuring that mappings between domains (e.g., A→B→A) remain consistent. In the field of natural language processing, back-translation employs similar ideas by translating sentences between languages in both directions to improve translation quality~\citep{sennrich2016backtranslation}. 
However, the cycle consistency in these works is in a single modality, \textit{i.e.} vision or language.
Recently, ITIT~\cite{li2023leveraging} was proposed to utilize cycle consistency to train vision-language generative models. ITIT takes in a mixture of unpaired data and paired data to pre-train the foundational generative model. It is constructed with one image-text encoder and two modality-specific
decoders, which operate on the encoded image-text features to generate either text or image tokens. In contrast, we utilized on single unified transformer to parse and predict text and image tokens together. Besides, we focus on adapting pre-trained foundational models to new domains efficiently rather than re-training new foundation models.

\begin{figure*}[t]
  \centering
    \includegraphics[width=1\linewidth]{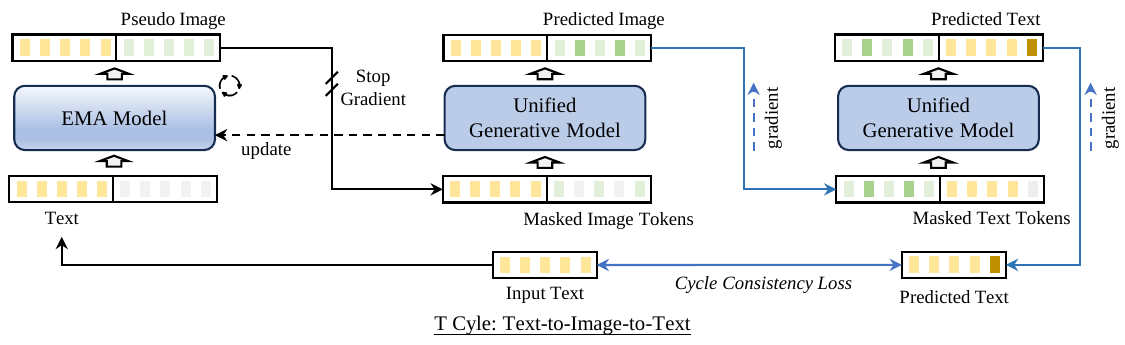}
   \caption{The overview of T cycle (text-to-image-to-text) of the proposed DoraCycle. The I cycle is similar and is omitted in the figure for brevity.}
   \label{fig:method}
\end{figure*}

\section{Method}

The proposed DoraCycle framework, as shown in Fig.~\ref{fig:method}, is built upon the unified generative model designed for multimodal tasks involving both vision and language~\cite{team2024chameleon, zhou2024transfusion, xie2024show}. The unified model uses a single transformer to learn bidirectional mappings between vision and language, providing a powerful backbone capable of processing and generating different modalities~\cite{xie2024show}. For captioning, the model takes in image tokens and predicts corresponding text tokens, while for image generation, it takes in text tokens and predicts image tokens. 
This versatility makes the unified model well-suited as a base for our proposed framework.

In the following sections, we first introduce the design of multimodal cycles, and then discuss the stabilization of optimization, and the balance of two cycles.

\subsection{Multimodal Cycles}
To adapt the unified model for domain-oriented adaptations using unpaired data, we design two multimodal cycles: the Image-to-Text-to-Image Cycle (I Cycle) and the Text-to-Image-to-Text Cycle (T Cycle). Each cycle utilizes data from a single modality, allowing the model to adapt without relying on paired data.

\textbf{T Cycle}: The T cycle training involves transforming an input textual sequence into an image representation and then back into a textual sequence, enforcing consistency between the generated and original text. Specifically, at each training iteration, we begin with an input text sequence $T = [t_l]_{l=1}^L$. Conditioned on this, the unified model generates pseudo-paired image tokens $I'$, representing the visual interpretation of the input text. The generated image tokens are then subjected to a random masking operation, denoted as $I'_M$, where a subset of the tokens is masked. The unified model is then called to reconstruct these masked tokens to form the complete synthesized image $\Tilde{I}$.

In the second half of the cycle, conditioning on image $\Tilde{I}$, the model generates the corresponding text sequence. The objective of the T cycle is to enforce cycle consistency between the generated text and the original input text $T$. The T cycle consistency loss is defined as follows:

\begin{equation}\label{t_cycle_loss}
\mathcal{L}_{TC} = -\mathbb{E}_{T \in \mathcal{D}{text}} \big[ \sum^{L}_{l=1}{ \log p(t_l | \Tilde{I}, t_0, \cdots, t_{l-1})} \big],
\end{equation}
where the $\mathcal{D}{text}$ is the set of text samples from the target domain.

\textbf{I Cycle}: 
The I cycle training begins with an input image, which is subsequently mapped to a textual representation and then transformed back to an image, enforcing consistency between the generated image and the original image tokens. At each training iteration, we start with input image tokens $I$. The unified model is used to synthesize pseudo-paired text tokens $T'$, representing the textual description of the image. We then use $T'$ in conjunction with the input image token $I$ to predict the reconstructed text tokens $\Tilde{T}$.

In the second half of the cycle, we pass the masked image tokens $I_M$ and the text $\Tilde{T}$ through the model to regenerate the masked image tokens. The cycle enforces consistency between the reconstructed and the original image tokens. The loss for enforcing I cycle consistency is given by:

\begin{equation}\label{i_cycle_loss}
\mathcal{L}_{IC} = -\mathbb{E}_{I \in \mathcal{D}_{image}} \big[ \sum_{\forall k: m_k=1} \log p(i_k | I_M, \Tilde{T}) \big],
\vspace{-5pt}
\end{equation}
where the $\mathcal{D}{image}$ is the set of image samples from the target domain, which are unpaired with the text samples.

By leveraging these two cycles, our framework forces the model to refine its generative understanding of both image and text representations, ensuring consistency between the input and output while effectively leveraging unpaired data to adapt the unified towards the target domain.

\textbf{Efficient Training}: In the intermediate steps of both cycles, generating the middle representation (i.e., captions or images) requires multiple forward passes. This is because the generation process involves either predicting the next tokens or the masked tokens multiple times. Backpropagating gradients through all these steps is computationally prohibitive. Thus, we first generate the intermediate results using the model in inference mode as pseudo-paired data, which are then used as the ground truth in the teacher-forcing scheme \cite{sutton1988learning, sutskever2014sequence} for the first half of the cycles. In this way, we reduce the number of forward passes to two, \textit{i.e.} one for generating the middle result and one for the final output, thus making the overall training process more memory efficient. 

\textbf{Token Differentiability}:
Since the intermediate outputs in each cycle are discrete tokens, which can not directly propagate gradients, we employ the Gumbel-Softmax\cite{jang2016categorical} to make these token representations differentiable.

\subsection{Stabilizing Optimization}
\label{sec:stable_train}
Each cycle involves the same unified model twice in the forward pass, which leads to optimization instabilities. To stabilize the training process, we adopt the Exponential Moving Average (EMA) training technique~\cite{tarvainen2017mean}. Specifically, we maintain a shadow version of the model, referred to as the EMA model, which is updated using an exponentially decaying average of the parameters of the main model.

\begin{equation}\label{ema_update}
\theta_{\text{EMA}} \leftarrow \alpha \theta_{\text{EMA}} + (1 - \alpha) \theta_{\text{main}},
\end{equation}
where $\alpha$ is a decay factor (set to 0.999) that controls the update rate, and $\theta_{\text{main}}$ represents the parameters of the main model.

In each training step, the EMA version of the model is used to generate the intermediate representation tokens (e.g., pseudo image or text tokens) which serve as pseudo ground truth during training. By using these stable targets from the slower-evolving EMA model, we can mitigate the risks of optimization instability. The main model is thus able to learn from more consistent and reliable intermediate targets, rather than being affected by fluctuations during the early stages of training.

\subsection{Balancing Two Cycles}

We observe that the T cycle tends to converge faster than the I cycle, primarily because textual data is inherently one-dimensional and simpler to learn compared to images. This imbalance in optimization leads to a kind of collapse of the model, where it tends to generate irrelevant but self-consistent captions for images, ultimately degrading the image-text alignment capability.

To address this problem, we make the gradients of the T cycle orthogonal to those of the I cycle, thus preventing interference. This is achieved by modifying the gradients using gradient surgery~\cite{yu2020gradient}.
Let $g_T$ and $g_I$ represent the gradients of the T cycle and the I cycle, respectively. We project $g_T$ onto the orthogonal complement of $g_I$ to obtain the modified gradient $g_T'$, which is defined as:

\begin{equation}
g_T' = g_T - \frac{g_T \cdot g_I}{g_I \cdot g_I} g_I,
\end{equation}
where $g_T \cdot g_I$ denotes the dot product between the gradients of the T and I cycle.

Additionally, we reweight the losses to further balance the learning between the I and T cycles. The final loss function is as,
\begin{equation}\label{i_cycle_loss}
\mathcal{L} = \mathcal{L}_{IC} + \beta \mathcal{L}_{TC},
\vspace{-5pt}
\end{equation}
where the $\beta$ is the weight of the T cycle loss.
\section{Experiments}
\subsection{Implementation Details}
To the best of our knowledge, Show-o~\cite{xie2024show} is currently the only fully open-source unified generative model with complete pre-trained weights and training code, including both its understanding and generation capabilities.
Therefore, we base DoraCycle on Show-o and conduct experiments accordingly. 
The base model is a unified transformer model that performs understanding and image generation by predicting discrete textual and visual tokens. 
We insert trainable low-rank adaptations (LoRA)~\cite{hu2022lora} modules into the Q projection and V projection of the attention layers from layers 7 to 24. The LoRA rank is set to 32. The $\beta$ is set to 0.1 to balance the optimization of two cycles.

The training of DoraCycle is performed on 8 NVIDIA H100 GPUs with mixed precision enabled for memory efficiency. 
We set the batch size to 32, with each cycle taking half of the batch when both cycles are being optimized simultaneously. 
The learning rate is set to $1e^{-4}$ with a cosine annealing schedule. 
The optimizer is AdamW with weight decay of $1e^{-2}$. Additionally, EMA is employed to stabilize the training process, as described in Section~\ref{sec:stable_train}.

\subsection{Domain-Oriented Adaptations}
\begin{figure}[t]  
    \centering
\includegraphics[width=1.0\linewidth]{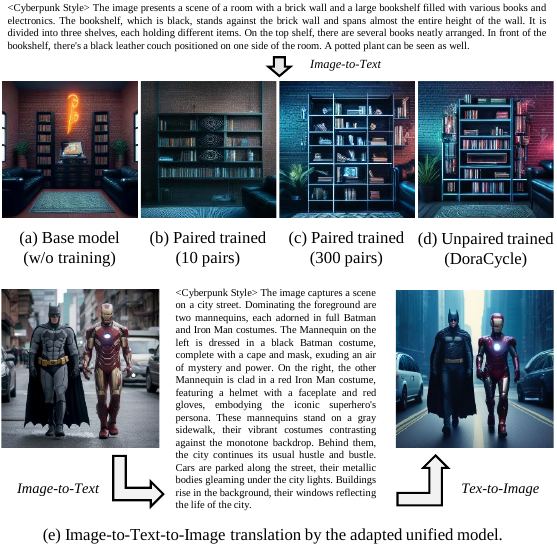}
    \caption{Domain-oriented adaptation with different training setups. (a) Image generated by the base model without training for adoption. (b) Image generated by the model trained with 10 paired image-text samples. (c) Image generated by the model trained with 300 paired image-text samples. (d) Image generated by the model trained by DoraCycle on only unpaired data. (e) Image-to-Text-to-Image translation performed by the adapted model trained by DoraCycle.}
\label{fig:results_1}
\end{figure}

\begin{figure*}[!tb]
    \centering
\includegraphics[width=0.98\linewidth]{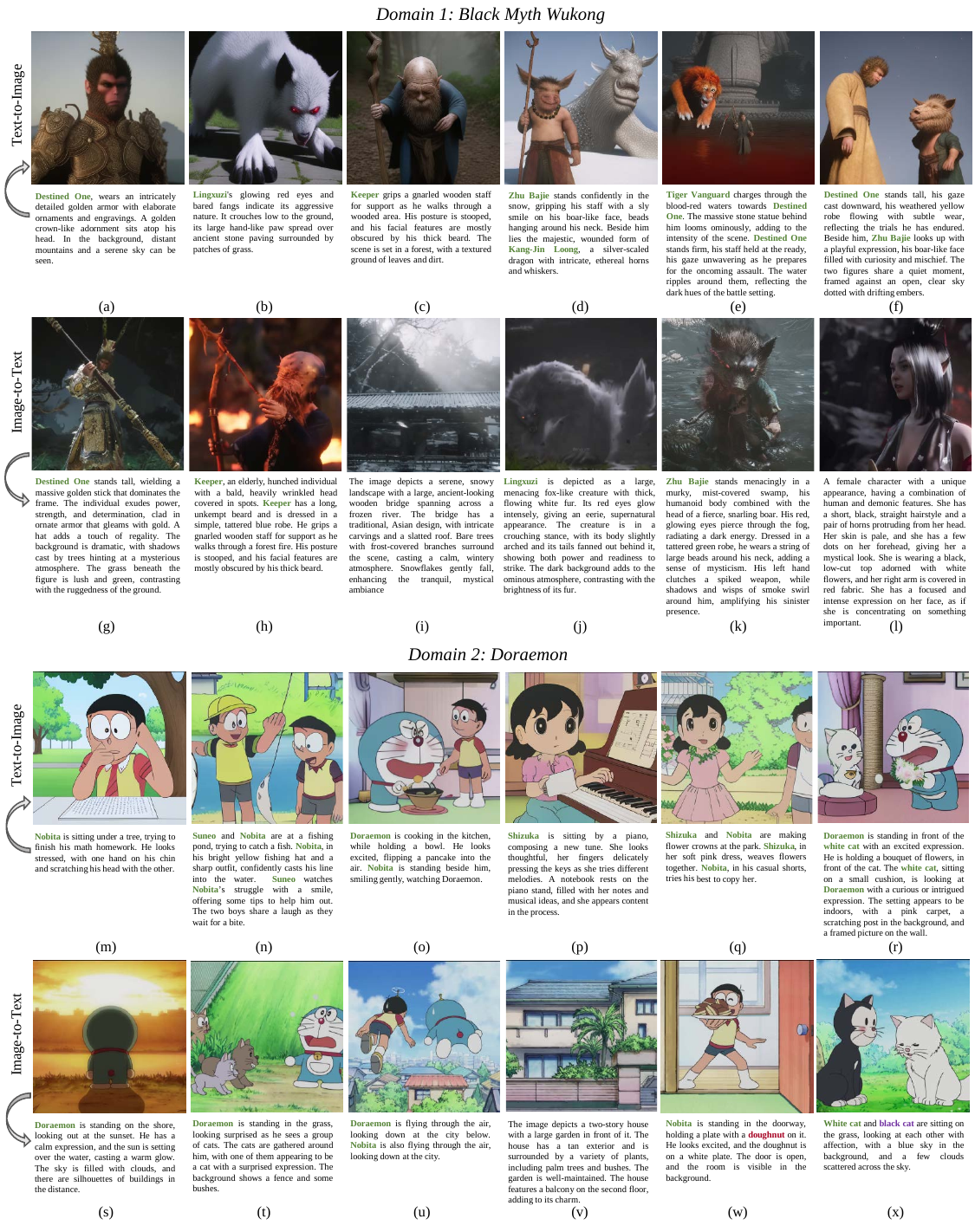}
    \caption{Image-to-text and text-to-image generation by the unified models that adapted for two domains. The special tokens are omitted.}
\label{fig:results_2}
\end{figure*}

\begin{figure}[t]  
    \centering
\includegraphics[width=1.0\linewidth]{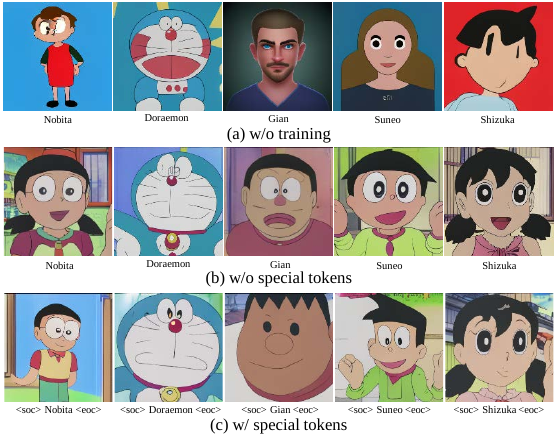}
    \caption{Effect of special tokens on character learning. (a) Base model without training. (b) Model trained without using special tokens, showing attribute confusion among characters. (c) Model trained with special tokens, improving character attribute alignment and reducing confusion.}
\label{fig:results_3}
\end{figure}

\textbf{Unpaired Training}: 
For tasks that do not require strongly related paired knowledge, our DoraCycle can fully learn the target domain using unpaired data. For example, to learn the cyberpunk style, we collected 300 cyberpunk-style images as input for the I cycle, and used the text data from the base model pre-training dataset~\citep{changpinyo2021conceptual} for the T cycle, with the keyword "cyberpunk style" automatically injected into text, prompting the model about the target style we want.

The experimental results are shown in Fig.~\ref{fig:results_1}. Given the same text prompt to generate cyberpunk-style images, Fig.~\ref{fig:results_1} (a) shows the images generated by the base model without additional training. It can be observed that the base model adds some cyberpunk elements, such as neon lights, but the overall atmosphere does not align well with the desired style. Fig.~\ref{fig:results_1} (d) shows the images generated by the adapted model trained with DoraCycle, which aligns well with the target style. Traditional text-to-image customization or adaptation methods, such as DreamBooth~\citep{ruiz2023dreambooth}, rely on paired data for training. Therefore, we simulate user-created paired data by annotating the collected images with captions, and split them into two groups. One group contained only 10 paired examples, which is an acceptable workload for users, while the other group contained captions for all 300 images, which would be labor-intensive and impractical for users. The images generated by the model that trained on 10 paired examples are shown in Fig.~\ref{fig:results_1} (b). It struggled to produce good stylized images, likely because the combination of indoor bookshelves with the cyberpunk style is too novel for the model to generalize well from limited paired data. The images generated by the model trained on 300 paired examples are shown in Fig.~\ref{fig:results_1} (c), which have better outputs. In contrast, the model trained using DoraCycle does not require manual captioning, significantly reducing the workload for users.

Fig.~\ref{fig:results_1} (e) illustrates the adapted model trained by DoraCyle maintains semantic consistency through image-to-text-to-image translation. The input image is transformed into a textual description and then reconstructed into an image. The result shows that the adapted model successfully captures and retains the key visual components in the original image throughout the multimodal cycle. 
Notably, the identity of the characters and the details of the environment are all preserved, indicating effective bidirectional understanding and generation capabilities in the target domain. 
Furthermore, the newly generated image incorporates styles learned from the target domain, demonstrating the generalization of the learned knowledge to images in the wild.

\textbf{Learning Paired Knowledge}: For tasks that require learning some paired knowledge, such as associating an identity name with its visual appearance, DoraCycle can incorporate a small amount of paired data to learn such associations while leveraging a large amount of unpaired data to comprehensively learn general aspects of the target domain. Specifically, in each batch, for data with paired ground truth, we compute the token prediction loss and also include it in the cycle, use ground truth as the pseudo middle generations, and compute the cycle loss. For unpaired data, we compute the unpaired cycle loss.

For example, when adapting the model to Domain 1: Black Myth Wukong and Domain 2: Doraemon, we annotate 1-3 images per unique identity with captions that specify the name of the identity. For each domain, we collect 2k images, which are mostly sampled from online videos, and independently collect text descriptions, which are further expanded to 1k by ChatGPT~\cite{ChatGPT2024}. The final adapted model trained with DoraCycle demonstrates strong performance in both text-to-image generation and image-to-text generation, as shown in Fig.~\ref{fig:results_2}.

In terms of text-to-image results, the model trained with DoraCycle effectively generated images that aligned well with the target domains. In domain 1 (Black Myth Wukong), the generated images accurately depicted domain-specific visual elements, such as the intricate details of character appearances and the overall fantasy-like atmosphere. This indicates that the model successfully learned to generalize the visual features from text prompts to realistic images within the target domain. Similarly, in domain 2 (Doraemon), the generated images preserve the iconic cartoonish aesthetics and capture key visual details of the characters and settings, demonstrating effective domain adaptation.

For the image-to-text task, the model performs well in generating contextually accurate captions. In domain 1, the generated captions provide rich descriptions of the characters, their attributes, and the context, effectively mirroring the visual elements present in the input images. In domain 2, the captions correctly describe the characters, their actions, and their environments concisely, maintaining consistency with the visual style. The ability of the model to generate accurate descriptions highlights its robust understanding of the visual components of the domain.

Additionally, an interesting phenomenon can be observed in how the model handles the visual elements that are not annotated with paired data. For instance, in Fig.~\ref{fig:results_2} (w), the dorayaki (a type of sweet bean-filled pancake) was described by the model as a "doughnut". This may be due to the fact that the anime-style representation of the dorayaki is novel, and neither the base model nor the unpaired training provided specific textual-visual pairing knowledge about it. On the other hand, in the example shown in Fig.~\ref{fig:results_2} (x), we annotate the white cat as a character with paired textual and visual data, using a special token for its name: "\texttt{<soc>} white cat \texttt{<eoc>}". Interestingly, although no paired annotation is provided for the black cat, the model still predicts the special token for it as "\texttt{<soc>} black cat \texttt{<eoc>}" during the caption generation. This suggests that the model autonomously categorized the black cat as a character when learning the target domain, indicating that it may have attempted to generalize learned knowledge from one type of entity to similar ones.

\begin{table*}[!tb]
\centering
\caption{Comparison of different training methods under various data settings. The best value is highlighted in \colorbox{pearDark!20}{blue}, and the second-best value is highlighted in \colorbox{mycolor_green}{green}. ``P" indicates paired data, and ``U" indicates unpaired data. } 
\label{tab:comp}{
\resizebox{1.\linewidth}{!}
{ 
\begin{tabular}
{>{\centering\arraybackslash}p{2.0cm}>{\centering\arraybackslash}p{1.0cm}>{\centering\arraybackslash}p{1.0cm}>{\centering\arraybackslash}p{2.3cm}>{\centering\arraybackslash}p{2.3cm} |>{\centering\arraybackslash}p{1.3cm}>{\centering\arraybackslash}p{1.3cm}>{\centering\arraybackslash}p{1.7cm}>{\centering\arraybackslash}p{1.7cm}}
\toprule
\multirow{2}{*}{ } & 
\multirow{2}{*}{T cycle} &  \multirow{2}{*}{I cycle} &  
\multirow{2}{*}{T Data} & \multirow{2}{*}{I Data} &  
\multirow{2}{*}{FID-1K $\downarrow$} &  \multirow{2}{*}{CIDEr $\uparrow$} &
 \multicolumn{2}{c}{Human Eval}  \\
&&&&&&  & T2I Align $\uparrow$ & I2T Align$\uparrow$ \\
\midrule
\multirow{2}{*}{DreamBooth~\cite{ruiz2023dreambooth}} & - & - & 10\% P & 10\% P& 33.22 &  32.74 & 3.25& 1.83\\
 & - & - & 100\% P & 100\% P& \colorbox{pearDark!20}{24.93}  &  \colorbox{pearDark!20}{41.55}  & \colorbox{pearDark!20}{4.13}
& \colorbox{pearDark!20}{3.96}\\ 
\midrule
\multirow{4}{*}{DoraCycle}& \ding{55} & \ding{51} & \ding{55} & 100\% U & 28.93 & 30.54 & 3.38 & 1.62\\ 
 & \ding{51}& \ding{55} & 100\% U  & \ding{55} & 36.63 & 35.70  & 3.26 & 2.17\\
 & \ding{51}& \ding{51} & 100\% U &  100\% U & 27.44 &  38.17 & 3.84 & 3.42\\
 & \ding{51}& \ding{51} & 10\% P + 90\% U &  10\% P + 90\% U & \colorbox{mycolor_green}{25.37} & \colorbox{mycolor_green}{40.90}  & \colorbox{mycolor_green}{4.12} & \colorbox{mycolor_green}{3.81}\\
\midrule
ITIT~\cite{li2023leveraging} & \ding{51}& \ding{51} & 10\% P + 90\% U &  10\% P + 90\% U & 27.50 &  38.62 & 3.85 & 3.52 \\
\bottomrule
\end{tabular}}}

\end{table*}

\textbf{Enhanced Learning with Special Tokens}:
As shown in Fig.~\ref{fig:results_3}, we experimentally find that the model often confused multiple novel concepts in the target domain. Fig.~\ref{fig:results_3} (a) shows the image generated by the base model without training taking in the name of characters. Fig.~\ref{fig:results_3} (b) shows the characters generated by the trained model. During training, the names of characters are directly included in the text without special treatment, leading to attribute confusion between characters. The varying lengths of the tokenized character names also make learning difficult. To solve this problem, we introduce a simple yet efficient solution: adding special tokens around character names. We introduced start of character (\texttt{<soc>}) and end of character (\texttt{<eoc>}) tokens to enclose character names, which significantly enhance the learning of novel concepts. As shown in Fig.~\ref{fig:results_3} (c), involving special tokens improves the alignments between characters and their names.

\subsection{Comparisons}
In this section, we use the Storyboard20K~\citep{xie2024learning} dataset to conduct the quantitative comparison experiments. The storyboards originating from the same data source are grouped to form a domain, consisting of images and descriptive text. The data are used under three different settings, \textit{i.e.} totally unpaired, only paired, and paired plus unpaired data, as shown in Table ~\ref{tab:comp}.

The compared methods include DreamBooth~\cite{ruiz2023dreambooth} and ITIT~\cite{li2023leveraging}. We implement DreamBooth as a paired-training baseline by applying LoRA fine-tuning on the unified model. The original design of ITIT is different, in which the image and text decoders are separate models, and its code has not been released. We adjusted and re-implemented it to be suitable for our unified model architecture.

We use both automatic and human evaluations to compare the performance of different methods. For automatic evaluation, we use FID to measure the distribution differences between the generated images and the target domain images~\cite{heusel2017gans}, and CIDEr to compute the error between the generated text and the ground truth~\cite{vedantam2015cider}. For human evaluation, we create 100 questions for the generated results of models, each rated by three different human raters. The raters are asked to evaluate the alignment between the image and text on a scale from 1 to 5, where 1 indicates no relevance and 5 indicates complete alignment.

The experimental results in Table~\ref{tab:comp} demonstrate that the proposed DoraCycle performs competitively under several data settings. Specifically, when using a combination of paired and unpaired data, DoraCycle outperforms ITIT. Compared to DreamBooth, which heavily relies on paired data, DoraCycle outperforms it when using the same scale of paired data, \textit{i.e.} 10\% paired data, indicting the benefits brought by 90\% unpaired data. While Dreambooth with 100\% achieves the best evaluation scores, the scores of the DoraCycle with 10\% paired and 90\% unpaired data are comparable with them.

Table~\ref{tab:comp} also shows the difference in the performance of DoraCyle under different cycle settings. It is shown that without the T cycle and with only the I cycle, the captioning ability of the adapted model degrades more significantly. In contrast, if only the T cycle is used and without the I cycle, the FID score increases substantially, indicating that the generated image distribution mismatches with the target distribution.

\begin{wraptable}{r}{0.25\textwidth}
\vspace{-5mm}
\centering
\caption{Ablation Studies. EMA refers to the exponential moving average. GS refers to gradient surgery.}
\label{tab:ablation}{
\resizebox{1.\linewidth}{!}
{ 
\begin{tabular}{l|cc}
\toprule
 &  FID-1K $\downarrow$ &  CIDEr $\uparrow$ \\
\midrule
w/o EMA & 27.19 & 38.85 \\
w/o GS & 25.54 & 39.98 \\
DoraCycle &  25.37 & 40.90 \\

\bottomrule
\end{tabular}}}
\vspace{-2mm}
\end{wraptable}

\subsection{Ablation Studies}
Table~\ref{tab:ablation} shows that removing key components from DoraCycle significantly impacts performance. Without EMA, the FID score increases from 25.37 to 27.19, indicating lower image quality due to less stable training. Removing Gradient Surgery (GS) will reduce the CIDEr score and increase the FID, indicating a worse performance. This demonstrates the importance of mitigating the interference between the optimization directions of two cycles. The complete DoraCycle framework, with both EMA and GS, has the best performance across all metrics, demonstrating the importance of these components in achieving better optimization.
\section{Conclusion}
We propose the DoraCycle to adapt the unified generative model to target domains within multimodal cycles. By leveraging both image-to-text-to-image and text-to-image-to-text cycles, DoraCycle changes the learning objectives into the same modality, allowing for effective optimization using unpaired data. Our experiments show that DoraCycle can adapt the unified model to target domains using only unpaired data, or involving a small amount of paired data when necessary to learn specific concepts. Experimental results demonstrate that DoraCycle achieves advanced or comparable performance across various settings. Leveraging unpaired data broadens the application potential of DoraCycle, making it ideally suited for domain adaptation tasks where paired data is scarce or challenging to collect.

{
    \small
    \bibliographystyle{ieeenat_fullname}
    \bibliography{main}

\begin{thebibliography}{87}
\providecommand{\natexlab}[1]{#1}
\providecommand{\url}[1]{\texttt{#1}}
\expandafter\ifx\csname urlstyle\endcsname\relax
  \providecommand{\doi}[1]{doi: #1}\else
  \providecommand{\doi}{doi: \begingroup \urlstyle{rm}\Url}\fi

\bibitem[Aghajanyan et~al.(2022)Aghajanyan, Huang, Ross, Karpukhin, Xu, Goyal, Okhonko, Joshi, Ghosh, Lewis, et~al.]{aghajanyan2022cm3}
Armen Aghajanyan, Bernie Huang, Candace Ross, Vladimir Karpukhin, Hu Xu, Naman Goyal, Dmytro Okhonko, Mandar Joshi, Gargi Ghosh, Mike Lewis, et~al.
\newblock Cm3: A causal masked multimodal model of the internet.
\newblock \emph{arXiv preprint arXiv:2201.07520}, 2022.

\bibitem[Aiello et~al.(2024)Aiello, YU, Nie, Aghajanyan, and Oguz]{jointly}
Emanuele Aiello, LILI YU, Yixin Nie, Armen Aghajanyan, and Barlas Oguz.
\newblock Jointly training large autoregressive multimodal models.
\newblock In \emph{ICLR}, 2024.

\bibitem[Balaji et~al.(2022)Balaji, Nah, Huang, Vahdat, Song, Kreis, Aittala, Aila, Laine, Catanzaro, et~al.]{balaji2022ediffi}
Yogesh Balaji, Seungjun Nah, Xun Huang, Arash Vahdat, Jiaming Song, Karsten Kreis, Miika Aittala, Timo Aila, Samuli Laine, Bryan Catanzaro, et~al.
\newblock ediffi: Text-to-image diffusion models with an ensemble of expert denoisers.
\newblock \emph{arXiv preprint arXiv:2211.01324}, 2022.

\bibitem[Changpinyo et~al.(2021)Changpinyo, Sharma, Ding, and Soricut]{changpinyo2021conceptual}
Soravit Changpinyo, Piyush Sharma, Nan Ding, and Radu Soricut.
\newblock Conceptual 12m: Pushing web-scale image-text pre-training to recognize long-tail visual concepts.
\newblock In \emph{Proceedings of the IEEE/CVF conference on computer vision and pattern recognition}, pages 3558--3568, 2021.

\bibitem[Chen et~al.(2024{\natexlab{a}})Chen, Ge, Xie, Wu, Yao, Ren, Wang, Luo, Lu, and Li]{chen2024pixart-sigma}
Junsong Chen, Chongjian Ge, Enze Xie, Yue Wu, Lewei Yao, Xiaozhe Ren, Zhongdao Wang, Ping Luo, Huchuan Lu, and Zhenguo Li.
\newblock Pixart-$\backslash$sigma: Weak-to-strong training of diffusion transformer for 4k text-to-image generation.
\newblock \emph{arXiv preprint arXiv:2403.04692}, 2024{\natexlab{a}}.

\bibitem[Chen et~al.(2024{\natexlab{b}})Chen, Yu, Ge, Yao, Xie, Wang, Kwok, Luo, Lu, and Li]{pixart}
Junsong Chen, Jincheng Yu, Chongjian Ge, Lewei Yao, Enze Xie, Zhongdao Wang, James~T. Kwok, Ping Luo, Huchuan Lu, and Zhenguo Li.
\newblock Pixart-{\(\alpha\)}: Fast training of diffusion transformer for photorealistic text-to-image synthesis.
\newblock In \emph{{ICLR}}. OpenReview.net, 2024{\natexlab{b}}.

\bibitem[Chen et~al.(2023)Chen, Zhao, Liu, Ding, Song, Wang, Wang, Yang, Liu, Du, et~al.]{chen2023photoverse}
Li Chen, Mengyi Zhao, Yiheng Liu, Mingxu Ding, Yangyang Song, Shizun Wang, Xu Wang, Hao Yang, Jing Liu, Kang Du, et~al.
\newblock Photoverse: Tuning-free image customization with text-to-image diffusion models.
\newblock \emph{arXiv preprint arXiv:2309.05793}, 2023.

\bibitem[Chen et~al.(2020)Chen, Radford, Child, Wu, Jun, Luan, and Sutskever]{chen2020generative}
Mark Chen, Alec Radford, Rewon Child, Jeffrey Wu, Heewoo Jun, David Luan, and Ilya Sutskever.
\newblock Generative pretraining from pixels.
\newblock In \emph{ICML}, pages 1691--1703, 2020.

\bibitem[Chen et~al.(2024{\natexlab{c}})Chen, Huang, Liu, Shen, Zhao, and Zhao]{chen2024anydoor}
Xi Chen, Lianghua Huang, Yu Liu, Yujun Shen, Deli Zhao, and Hengshuang Zhao.
\newblock Anydoor: Zero-shot object-level image customization.
\newblock In \emph{Proceedings of the IEEE/CVF Conference on Computer Vision and Pattern Recognition}, pages 6593--6602, 2024{\natexlab{c}}.

\bibitem[Choi et~al.(2018)Choi, Choi, Kim, Ha, Kim, and Choo]{choi2018stargan}
Yunjey Choi, Minje Choi, Munyoung Kim, Jung-Woo Ha, Sunghun Kim, and Jaegul Choo.
\newblock Stargan: Unified generative adversarial networks for multi-domain image-to-image translation.
\newblock In \emph{Proceedings of the IEEE Conference on Computer Vision and Pattern Recognition (CVPR)}, 2018.

\bibitem[Dai et~al.(2023{\natexlab{a}})Dai, Li, Li, Tiong, Zhao, Wang, Li, Fung, and Hoi]{instructblip}
Wenliang Dai, Junnan Li, Dongxu Li, Anthony Meng~Huat Tiong, Junqi Zhao, Weisheng Wang, Boyang Li, Pascale Fung, and Steven Hoi.
\newblock Instructblip: Towards general-purpose vision-language models with instruction tuning, 2023{\natexlab{a}}.

\bibitem[Dai et~al.(2023{\natexlab{b}})Dai, Hou, Ma, Tsai, Wang, Wang, Zhang, Vandenhende, Wang, Dubey, et~al.]{dai2023emu}
Xiaoliang Dai, Ji Hou, Chih-Yao Ma, Sam Tsai, Jialiang Wang, Rui Wang, Peizhao Zhang, Simon Vandenhende, Xiaofang Wang, Abhimanyu Dubey, et~al.
\newblock Emu: Enhancing image generation models using photogenic needles in a haystack.
\newblock \emph{arXiv preprint arXiv:2309.15807}, 2023{\natexlab{b}}.

\bibitem[Dehghani et~al.(2023)Dehghani, Djolonga, Mustafa, Padlewski, Heek, Gilmer, Steiner, Caron, Geirhos, Alabdulmohsin, Jenatton, Beyer, Tschannen, Arnab, Wang, Ruiz, Minderer, Puigcerver, Evci, Kumar, van Steenkiste, Elsayed, Mahendran, Yu, Oliver, Huot, Bastings, Collier, Gritsenko, Birodkar, Vasconcelos, Tay, Mensink, Kolesnikov, Pavetic, Tran, Kipf, Lucic, Zhai, Keysers, Harmsen, and Houlsby]{qknorm}
Mostafa Dehghani, Josip Djolonga, Basil Mustafa, Piotr Padlewski, Jonathan Heek, Justin Gilmer, Andreas~Peter Steiner, Mathilde Caron, Robert Geirhos, Ibrahim Alabdulmohsin, Rodolphe Jenatton, Lucas Beyer, Michael Tschannen, Anurag Arnab, Xiao Wang, Carlos~Riquelme Ruiz, Matthias Minderer, Joan Puigcerver, Utku Evci, Manoj Kumar, Sjoerd van Steenkiste, Gamaleldin~Fathy Elsayed, Aravindh Mahendran, Fisher Yu, Avital Oliver, Fantine Huot, Jasmijn Bastings, Mark Collier, Alexey~A. Gritsenko, Vighnesh Birodkar, Cristina~Nader Vasconcelos, Yi Tay, Thomas Mensink, Alexander Kolesnikov, Filip Pavetic, Dustin Tran, Thomas Kipf, Mario Lucic, Xiaohua Zhai, Daniel Keysers, Jeremiah~J. Harmsen, and Neil Houlsby.
\newblock Scaling vision transformers to 22 billion parameters.
\newblock In \emph{ICML}, pages 7480--7512, 2023.

\bibitem[Dhariwal and Nichol(2021)]{dhariwal2021diffusion}
Prafulla Dhariwal and Alexander Nichol.
\newblock Diffusion models beat gans on image synthesis.
\newblock \emph{Advances in Neural Information Processing Systems}, 34:\penalty0 8780--8794, 2021.

\bibitem[Dong et~al.(2024)Dong, Han, Peng, Qi, Ge, Yang, Zhao, Sun, Zhou, Wei, Kong, Zhang, Ma, and Yi]{dreamllm}
Runpei Dong, Chunrui Han, Yuang Peng, Zekun Qi, Zheng Ge, Jinrong Yang, Liang Zhao, Jianjian Sun, Hongyu Zhou, Haoran Wei, Xiangwen Kong, Xiangyu Zhang, Kaisheng Ma, and Li Yi.
\newblock Dream{LLM}: Synergistic multimodal comprehension and creation.
\newblock In \emph{ICLR}, 2024.

\bibitem[Dwibedi et~al.(2019)Dwibedi, Aytar, Tompson, Sermanet, and Zisserman]{dwibedi2019temporal}
Debidatta Dwibedi, Yusuf Aytar, Jonathan Tompson, Pierre Sermanet, and Andrew Zisserman.
\newblock Temporal cycle-consistency learning.
\newblock In \emph{Proceedings of the IEEE/CVF conference on computer vision and pattern recognition}, pages 1801--1810, 2019.

\bibitem[Esser et~al.(2024)Esser, Kulal, Blattmann, Entezari, M{\"u}ller, Saini, Levi, Lorenz, Sauer, Boesel, et~al.]{esser2024scaling}
Patrick Esser, Sumith Kulal, Andreas Blattmann, Rahim Entezari, Jonas M{\"u}ller, Harry Saini, Yam Levi, Dominik Lorenz, Axel Sauer, Frederic Boesel, et~al.
\newblock Scaling rectified flow transformers for high-resolution image synthesis.
\newblock \emph{arXiv preprint arXiv:2403.03206}, 2024.

\bibitem[Gal et~al.(2022)Gal, Alaluf, Atzmon, Patashnik, Bermano, Chechik, and Cohen-Or]{gal2022image}
Rinon Gal, Yuval Alaluf, Yuval Atzmon, Or Patashnik, Amit~H Bermano, Gal Chechik, and Daniel Cohen-Or.
\newblock An image is worth one word: Personalizing text-to-image generation using textual inversion.
\newblock \emph{arXiv preprint arXiv:2208.01618}, 2022.

\bibitem[Ge et~al.(2024)Ge, Zhao, Zhu, Ge, Yi, Song, Li, Ding, and Shan]{seed-x}
Yuying Ge, Sijie Zhao, Jinguo Zhu, Yixiao Ge, Kun Yi, Lin Song, Chen Li, Xiaohan Ding, and Ying Shan.
\newblock Seed-x: Multimodal models with unified multi-granularity comprehension and generation.
\newblock \emph{arXiv preprint arXiv:2404.14396}, 2024.

\bibitem[Gu et~al.(2023)Gu, Wang, Wu, Shi, Chen, Fan, Xiao, Zhao, Chang, Wu, et~al.]{gu2023mix}
Yuchao Gu, Xintao Wang, Jay~Zhangjie Wu, Yujun Shi, Yunpeng Chen, Zihan Fan, Wuyou Xiao, Rui Zhao, Shuning Chang, Weijia Wu, et~al.
\newblock Mix-of-show: Decentralized low-rank adaptation for multi-concept customization of diffusion models.
\newblock \emph{Advances in Neural Information Processing Systems}, 36:\penalty0 15890--15902, 2023.

\bibitem[Guo et~al.(2023)Guo, Wang, Wu, Zhang, Wang, Xu, Song, Shi, and Huang]{guo2023zero}
Jiayi Guo, Chaofei Wang, You Wu, Eric Zhang, Kai Wang, Xingqian Xu, Shiji Song, Humphrey Shi, and Gao Huang.
\newblock Zero-shot generative model adaptation via image-specific prompt learning.
\newblock In \emph{Proceedings of the IEEE/CVF conference on computer vision and pattern recognition}, pages 11494--11503, 2023.

\bibitem[He et~al.(2016)He, Xia, Qin, Wang, Yu, Liu, and Ma]{he2016dual}
Di He, Yingce Xia, Tao Qin, Liwei Wang, Nenghai Yu, Tie-Yan Liu, and Wei-Ying Ma.
\newblock Dual learning for machine translation.
\newblock In \emph{Advances in Neural Information Processing Systems (NeurIPS)}, 2016.

\bibitem[Hessel et~al.(2023)Hessel, Marasovi{\'c}, Hwang, Lee, Da, Zellers, Mankoff, and Choi]{hessel2023androids}
Jack Hessel, Ana Marasovi{\'c}, Jena~D. Hwang, Lillian Lee, Jeff Da, Rowan Zellers, Robert Mankoff, and Yejin Choi.
\newblock Do androids laugh at electric sheep? {Humor} ``understanding'' benchmarks from {The New Yorker Caption Contest}.
\newblock In \emph{Proceedings of the ACL}, 2023.

\bibitem[Heusel et~al.(2017)Heusel, Ramsauer, Unterthiner, Nessler, and Hochreiter]{heusel2017gans}
Martin Heusel, Hubert Ramsauer, Thomas Unterthiner, Bernhard Nessler, and Sepp Hochreiter.
\newblock Gans trained by a two time-scale update rule converge to a local nash equilibrium.
\newblock In \emph{Proceedings of the 31st International Conference on Neural Information Processing Systems}, pages 6626--6637, 2017.

\bibitem[Hoffman et~al.(2018)Hoffman, Tzeng, Park, Zhu, Isola, Saenko, Efros, and Darrell]{hoffman2018cycada}
Judy Hoffman, Eric Tzeng, Taesung Park, Jun-Yan Zhu, Phillip Isola, Kate Saenko, Alexei~A Efros, and Trevor Darrell.
\newblock Cycada: Cycle-consistent adversarial domain adaptation.
\newblock In \emph{Proceedings of the International Conference on Machine Learning (ICML)}, 2018.

\bibitem[Hossain et~al.(2019)Hossain, Sohel, Shiratuddin, and Laga]{hossain2019comprehensive}
MD~Zakir Hossain, Ferdous Sohel, Mohd~Fairuz Shiratuddin, and Hamid Laga.
\newblock A comprehensive survey of deep learning for image captioning.
\newblock \emph{ACM Computing Surveys (CsUR)}, 51\penalty0 (6):\penalty0 1--36, 2019.

\bibitem[Hu et~al.(2022{\natexlab{a}})Hu, Shen, Wallis, Allen-Zhu, Li, Wang, and Chen]{hu2022lora}
Edward~J. Hu, Yelong Shen, Phillip Wallis, Zeyuan Allen-Zhu, Yuanzhi Li, Shean Wang, and Weizhu Chen.
\newblock Lora: Low-rank adaptation of large language models.
\newblock In \emph{International Conference on Learning Representations}, 2022{\natexlab{a}}.

\bibitem[Hu et~al.(2023)Hu, Cavicchioli, and Capotondi]{hu2023exploiting}
Jia~Cheng Hu, Roberto Cavicchioli, and Alessandro Capotondi.
\newblock Exploiting multiple sequence lengths in fast end to end training for image captioning.
\newblock In \emph{2023 IEEE International Conference on Big Data (BigData)}, pages 2173--2182. IEEE, 2023.

\bibitem[Hu et~al.(2022{\natexlab{b}})Hu, Gan, Wang, Yang, Liu, Lu, and Wang]{hu2022scaling}
Xiaowei Hu, Zhe Gan, Jianfeng Wang, Zhengyuan Yang, Zicheng Liu, Yumao Lu, and Lijuan Wang.
\newblock Scaling up vision-language pre-training for image captioning.
\newblock In \emph{Proceedings of the IEEE/CVF conference on computer vision and pattern recognition}, pages 17980--17989, 2022{\natexlab{b}}.

\bibitem[Jain et~al.(2020)Jain, Jamieson, Mankoff, Nowak, and Sievert]{newyorkernextmldataset}
Lalit Jain, Kevin Jamieson, Robert Mankoff, Robert Nowak, and Scott Sievert.
\newblock The {N}ew {Y}orker cartoon caption contest dataset, 2020.

\bibitem[Jang et~al.(2016)Jang, Gu, and Poole]{jang2016categorical}
Eric Jang, Shixiang Gu, and Ben Poole.
\newblock Categorical reparameterization with gumbel-softmax.
\newblock \emph{arXiv preprint arXiv:1611.01144}, 2016.

\bibitem[Jia et~al.(2023)Jia, Zhao, Chan, Li, Zhang, Gong, Hou, Wang, and Su]{jia2023taming}
Xuhui Jia, Yang Zhao, Kelvin~CK Chan, Yandong Li, Han Zhang, Boqing Gong, Tingbo Hou, Huisheng Wang, and Yu-Chuan Su.
\newblock Taming encoder for zero fine-tuning image customization with text-to-image diffusion models.
\newblock \emph{arXiv preprint arXiv:2304.02642}, 2023.

\bibitem[Kumari et~al.(2023)Kumari, Zhang, Zhang, Shechtman, and Zhu]{kumari2023multi}
Nupur Kumari, Bingliang Zhang, Richard Zhang, Eli Shechtman, and Jun-Yan Zhu.
\newblock Multi-concept customization of text-to-image diffusion.
\newblock In \emph{Proceedings of the IEEE/CVF Conference on Computer Vision and Pattern Recognition}, pages 1931--1941, 2023.

\bibitem[Li et~al.(2022)Li, Xu, Tian, Wang, Yan, Bi, Ye, Chen, Xu, Cao, et~al.]{li2022mplug}
Chenliang Li, Haiyang Xu, Junfeng Tian, Wei Wang, Ming Yan, Bin Bi, Jiabo Ye, Hehong Chen, Guohai Xu, Zheng Cao, et~al.
\newblock mplug: Effective and efficient vision-language learning by cross-modal skip-connections.
\newblock \emph{arXiv preprint arXiv:2205.12005}, 2022.

\bibitem[Li et~al.(2024)Li, Gan, Yang, Yang, Li, Wang, Gao, et~al.]{li2024multimodal}
Chunyuan Li, Zhe Gan, Zhengyuan Yang, Jianwei Yang, Linjie Li, Lijuan Wang, Jianfeng Gao, et~al.
\newblock Multimodal foundation models: From specialists to general-purpose assistants.
\newblock \emph{Foundations and Trends{\textregistered} in Computer Graphics and Vision}, 16\penalty0 (1-2):\penalty0 1--214, 2024.

\bibitem[Li et~al.(2023{\natexlab{a}})Li, Li, Savarese, and Hoi]{li2023blip}
Junnan Li, Dongxu Li, Silvio Savarese, and Steven Hoi.
\newblock Blip-2: Bootstrapping language-image pre-training with frozen image encoders and large language models.
\newblock In \emph{International conference on machine learning}, pages 19730--19742. PMLR, 2023{\natexlab{a}}.

\bibitem[Li et~al.(2023{\natexlab{b}})Li, Bhardwaj, Tian, Zhang, Barber, Katabi, Lajoie, Chang, and Krishnan]{li2023leveraging}
Tianhong Li, Sangnie Bhardwaj, Yonglong Tian, Han Zhang, Jarred Barber, Dina Katabi, Guillaume Lajoie, Huiwen Chang, and Dilip Krishnan.
\newblock Leveraging unpaired data for vision-language generative models via cycle consistency.
\newblock \emph{arXiv preprint arXiv:2310.03734}, 2023{\natexlab{b}}.

\bibitem[Li et~al.(2023{\natexlab{c}})Li, Bubeck, Eldan, Del~Giorno, Gunasekar, and Lee]{phi1.5}
Yuanzhi Li, S{\'e}bastien Bubeck, Ronen Eldan, Allie Del~Giorno, Suriya Gunasekar, and Yin~Tat Lee.
\newblock Textbooks are all you need ii: phi-1.5 technical report.
\newblock \emph{arXiv preprint arXiv:2309.05463}, 2023{\natexlab{c}}.

\bibitem[Liu et~al.(2024)Liu, Li, Wu, and Lee]{llava}
Haotian Liu, Chunyuan Li, Qingyang Wu, and Yong~Jae Lee.
\newblock Visual instruction tuning.
\newblock \emph{NeurIPS}, 36, 2024.

\bibitem[Ma et~al.(2024{\natexlab{a}})Ma, Goldstein, Albergo, Boffi, Vanden-Eijnden, and Xie]{ma2024sit}
Nanye Ma, Mark Goldstein, Michael~S Albergo, Nicholas~M Boffi, Eric Vanden-Eijnden, and Saining Xie.
\newblock Sit: Exploring flow and diffusion-based generative models with scalable interpolant transformers.
\newblock \emph{arXiv preprint arXiv:2401.08740}, 2024{\natexlab{a}}.

\bibitem[Ma et~al.(2024{\natexlab{b}})Ma, Wang, Jia, Chen, Liu, Li, Chen, and Qiao]{ma2024latte}
Xin Ma, Yaohui Wang, Gengyun Jia, Xinyuan Chen, Ziwei Liu, Yuan-Fang Li, Cunjian Chen, and Yu Qiao.
\newblock Latte: Latent diffusion transformer for video generation.
\newblock \emph{arXiv preprint arXiv:2401.03048}, 2024{\natexlab{b}}.

\bibitem[Mokady et~al.(2023)Mokady, Hertz, Aberman, Pritch, and Cohen-Or]{mokady2023null}
Ron Mokady, Amir Hertz, Kfir Aberman, Yael Pritch, and Daniel Cohen-Or.
\newblock Null-text inversion for editing real images using guided diffusion models.
\newblock In \emph{Proceedings of the IEEE/CVF Conference on Computer Vision and Pattern Recognition}, pages 6038--6047, 2023.

\bibitem[{OpenAI}(2024)]{ChatGPT2024}
{OpenAI}.
\newblock Chatgpt.
\newblock \url{https://chatgpt.com/}, 2024.

\bibitem[Radev et~al.(2016)Radev, Stent, Tetreault, Pappu, Iliakopoulou, Chanfreau, de~Juan, Vallmitjana, Jaimes, Jha, and Mankoff]{radev-etal-2016-humor}
Dragomir Radev, Amanda Stent, Joel Tetreault, Aasish Pappu, Aikaterini Iliakopoulou, Agustin Chanfreau, Paloma de Juan, Jordi Vallmitjana, Alejandro Jaimes, Rahul Jha, and Robert Mankoff.
\newblock Humor in collective discourse: Unsupervised funniness detection in the {New Yorker} cartoon caption contest.
\newblock In \emph{LREC}, 2016.

\bibitem[Radford et~al.(2021)Radford, Kim, Hallacy, Ramesh, Goh, Agarwal, Sastry, Askell, Mishkin, Clark, et~al.]{radford2021learning}
Alec Radford, Jong~Wook Kim, Chris Hallacy, Aditya Ramesh, Gabriel Goh, Sandhini Agarwal, Girish Sastry, Amanda Askell, Pamela Mishkin, Jack Clark, et~al.
\newblock Learning transferable visual models from natural language supervision.
\newblock In \emph{International conference on machine learning}, pages 8748--8763. PMLR, 2021.

\bibitem[Ramesh et~al.(2021)Ramesh, Pavlov, Goh, Gray, Voss, Radford, Chen, and Sutskever]{dalle}
Aditya Ramesh, Mikhail Pavlov, Gabriel Goh, Scott Gray, Chelsea Voss, Alec Radford, Mark Chen, and Ilya Sutskever.
\newblock Zero-shot text-to-image generation.
\newblock In \emph{ICML}, pages 8821--8831. Pmlr, 2021.

\bibitem[Ramesh et~al.(2022{\natexlab{a}})Ramesh, Dhariwal, Nichol, Chu, and Chen]{dalle2}
Aditya Ramesh, Prafulla Dhariwal, Alex Nichol, Casey Chu, and Mark Chen.
\newblock Hierarchical text-conditional image generation with {CLIP} latents.
\newblock \emph{CoRR}, abs/2204.06125, 2022{\natexlab{a}}.

\bibitem[Ramesh et~al.(2022{\natexlab{b}})Ramesh, Dhariwal, Nichol, Chu, and Chen]{ramesh2022dalle2}
Aditya Ramesh, Prafulla Dhariwal, Alex Nichol, Casey Chu, and Mark Chen.
\newblock Hierarchical diffusion models for text-to-image generation.
\newblock \emph{arXiv preprint arXiv:2204.06125}, 2022{\natexlab{b}}.

\bibitem[Ramesh et~al.(2022{\natexlab{c}})Ramesh, Dhariwal, Nichol, Chu, and Chen]{ramesh2022hierarchical}
Aditya Ramesh, Prafulla Dhariwal, Alex Nichol, Casey Chu, and Mark Chen.
\newblock Hierarchical text-conditional image generation with clip latents.
\newblock \emph{arXiv preprint arXiv:2204.06125}, 1\penalty0 (2):\penalty0 3, 2022{\natexlab{c}}.

\bibitem[Rombach et~al.(2022{\natexlab{a}})Rombach, Blattmann, Lorenz, Esser, and Ommer]{rombach2022high}
Robin Rombach, Andreas Blattmann, Dominik Lorenz, Patrick Esser, and Bj{\"o}rn Ommer.
\newblock High-resolution image synthesis with latent diffusion models.
\newblock In \emph{CVPR}, pages 10684--10695, 2022{\natexlab{a}}.

\bibitem[Rombach et~al.(2022{\natexlab{b}})Rombach, Blattmann, Lorenz, Esser, and Ommer]{rombach2022highresolution}
Robin Rombach, Andreas Blattmann, Dominik Lorenz, Patrick Esser, and Bj{\"o}rn Ommer.
\newblock High-resolution image synthesis with latent diffusion models.
\newblock \emph{arXiv preprint arXiv:2112.10752}, 2022{\natexlab{b}}.

\bibitem[Ruiz et~al.(2022)Ruiz, Li, Jampani, Pritch, Rubinstein, and Aberman]{ruiz2022dreambooth}
Nataniel Ruiz, Yuanzhen Li, Varun Jampani, Yael Pritch, Michael Rubinstein, and Kfir Aberman.
\newblock Dreambooth: Fine tuning text-to-image diffusion models for subject-driven generation.
\newblock \emph{arXiv preprint arXiv:2208.12242}, 2022.

\bibitem[Ruiz et~al.(2023)Ruiz, Li, Jampani, Pritch, Rubinstein, and Aberman]{ruiz2023dreambooth}
Nataniel Ruiz, Yuanzhen Li, Varun Jampani, Yael Pritch, Michael Rubinstein, and Kfir Aberman.
\newblock Dreambooth: Fine tuning text-to-image diffusion models for subject-driven generation.
\newblock In \emph{Proceedings of the IEEE/CVF conference on computer vision and pattern recognition}, pages 22500--22510, 2023.

\bibitem[Saharia et~al.(2022)Saharia, Chan, Saxena, Li, Whang, Denton, Ghasemipour, Gontijo~Lopes, Karagol~Ayan, Salimans, et~al.]{saharia2022photorealistic}
Chitwan Saharia, William Chan, Saurabh Saxena, Lala Li, Jay Whang, Emily~L Denton, Kamyar Ghasemipour, Raphael Gontijo~Lopes, Burcu Karagol~Ayan, Tim Salimans, et~al.
\newblock Photorealistic text-to-image diffusion models with deep language understanding.
\newblock \emph{Advances in Neural Information Processing Systems}, 35:\penalty0 36479--36494, 2022.

\bibitem[Sennrich et~al.(2016)Sennrich, Haddow, and Birch]{sennrich2016backtranslation}
Rico Sennrich, Barry Haddow, and Alexandra Birch.
\newblock Improving neural machine translation models with monolingual data.
\newblock In \emph{Proceedings of the 54th Annual Meeting of the Association for Computational Linguistics (Volume 1: Long Papers)}, 2016.

\bibitem[Shah et~al.(2019{\natexlab{a}})Shah, Chen, Rohrbach, and Parikh]{shah2019cycle}
Meet Shah, Xinlei Chen, Marcus Rohrbach, and Devi Parikh.
\newblock Cycle-consistency for robust visual question answering.
\newblock In \emph{Proceedings of the IEEE/CVF Conference on Computer Vision and Pattern Recognition}, pages 6649--6658, 2019{\natexlab{a}}.

\bibitem[Shah et~al.(2019{\natexlab{b}})Shah, Bharadwaj, Parikh, and Batra]{shah2019cyclevqa}
Sameer Shah, Siddharth Bharadwaj, Devi Parikh, and Dhruv Batra.
\newblock Cycle-consistency for robust visual question answering.
\newblock In \emph{Proceedings of the IEEE Conference on Computer Vision and Pattern Recognition (CVPR)}, 2019{\natexlab{b}}.

\bibitem[Shahaf et~al.(2015)Shahaf, Horvitz, and Mankoff]{shahaf2015inside}
Dafna Shahaf, Eric Horvitz, and Robert Mankoff.
\newblock Inside jokes: Identifying humorous cartoon captions.
\newblock In \emph{KDD}, 2015.

\bibitem[Sun et~al.(2024)Sun, Cui, Zhang, Zhang, Yu, Wang, Rao, Liu, Huang, and Wang]{sun2024generative}
Quan Sun, Yufeng Cui, Xiaosong Zhang, Fan Zhang, Qiying Yu, Yueze Wang, Yongming Rao, Jingjing Liu, Tiejun Huang, and Xinlong Wang.
\newblock Generative multimodal models are in-context learners.
\newblock In \emph{Proceedings of the IEEE/CVF Conference on Computer Vision and Pattern Recognition}, pages 14398--14409, 2024.

\bibitem[Sutskever(2014)]{sutskever2014sequence}
I Sutskever.
\newblock Sequence to sequence learning with neural networks.
\newblock \emph{arXiv preprint arXiv:1409.3215}, 2014.

\bibitem[Sutton(1988)]{sutton1988learning}
Richard~S Sutton.
\newblock Learning to predict by the methods of temporal differences.
\newblock \emph{Machine learning}, 3:\penalty0 9--44, 1988.

\bibitem[Tang et~al.(2024)Tang, Yang, Zhu, Zeng, and Bansal]{CoDI}
Zineng Tang, Ziyi Yang, Chenguang Zhu, Michael Zeng, and Mohit Bansal.
\newblock Any-to-any generation via composable diffusion.
\newblock \emph{NeurIPS}, 36, 2024.

\bibitem[Tarvainen and Valpola(2017)]{tarvainen2017mean}
Antti Tarvainen and Harri Valpola.
\newblock Mean teachers are better role models: Weight-averaged consistency targets improve semi-supervised deep learning results.
\newblock In \emph{Advances in Neural Information Processing Systems}, pages 1195--1204, 2017.

\bibitem[Team(2024)]{team2024chameleon}
Chameleon Team.
\newblock Chameleon: Mixed-modal early-fusion foundation models.
\newblock \emph{arXiv preprint arXiv:2405.09818}, 2024.

\bibitem[Vedantam et~al.(2015)Vedantam, Zitnick, and Parikh]{vedantam2015cider}
Ramakrishna Vedantam, C.~Lawrence Zitnick, and Devi Parikh.
\newblock Cider: Consensus-based image description evaluation.
\newblock In \emph{Proceedings of the IEEE Conference on Computer Vision and Pattern Recognition (CVPR)}, pages 4566--4575, 2015.

\bibitem[Wang et~al.(2022{\natexlab{a}})Wang, Yang, Hu, Li, Lin, Gan, Liu, Liu, and Wang]{wang2022git}
Jianfeng Wang, Zhengyuan Yang, Xiaowei Hu, Linjie Li, Kevin Lin, Zhe Gan, Zicheng Liu, Ce Liu, and Lijuan Wang.
\newblock Git: A generative image-to-text transformer for vision and language.
\newblock \emph{arXiv preprint arXiv:2205.14100}, 2022{\natexlab{a}}.

\bibitem[Wang et~al.(2022{\natexlab{b}})Wang, Yang, Men, Lin, Bai, Li, Ma, Zhou, Zhou, and Yang]{wang2022ofa}
Peng Wang, An Yang, Rui Men, Junyang Lin, Shuai Bai, Zhikang Li, Jianxin Ma, Chang Zhou, Jingren Zhou, and Hongxia Yang.
\newblock Ofa: Unifying architectures, tasks, and modalities through a simple sequence-to-sequence learning framework.
\newblock In \emph{International conference on machine learning}, pages 23318--23340. PMLR, 2022{\natexlab{b}}.

\bibitem[Wang et~al.(2024{\natexlab{a}})Wang, Bai, Wang, Qin, Chen, Li, Tang, and Hu]{wang2024instantid}
Qixun Wang, Xu Bai, Haofan Wang, Zekui Qin, Anthony Chen, Huaxia Li, Xu Tang, and Yao Hu.
\newblock Instantid: Zero-shot identity-preserving generation in seconds.
\newblock \emph{arXiv preprint arXiv:2401.07519}, 2024{\natexlab{a}}.

\bibitem[Wang et~al.(2024{\natexlab{b}})Wang, Ren, Gao, Yao, Guo, An, Bai, and Sun]{wang2024ladic}
Yuchi Wang, Shuhuai Ren, Rundong Gao, Linli Yao, Qingyan Guo, Kaikai An, Jianhong Bai, and Xu Sun.
\newblock Ladic: Are diffusion models really inferior to autoregressive counterparts for image-to-text generation?
\newblock \emph{arXiv preprint arXiv:2404.10763}, 2024{\natexlab{b}}.

\bibitem[Wortsman et~al.(2023)Wortsman, Liu, Xiao, Everett, Alemi, Adlam, Co-Reyes, Gur, Kumar, Novak, et~al.]{qknorm2}
Mitchell Wortsman, Peter~J Liu, Lechao Xiao, Katie Everett, Alex Alemi, Ben Adlam, John~D Co-Reyes, Izzeddin Gur, Abhishek Kumar, Roman Novak, et~al.
\newblock Small-scale proxies for large-scale transformer training instabilities.
\newblock \emph{arXiv preprint arXiv:2309.14322}, 2023.

\bibitem[Wu et~al.(2024)Wu, Chen, Wu, Ma, Liu, Pan, Liu, Xie, Yu, Ruan, et~al.]{wu2024janus}
Chengyue Wu, Xiaokang Chen, Zhiyu Wu, Yiyang Ma, Xingchao Liu, Zizheng Pan, Wen Liu, Zhenda Xie, Xingkai Yu, Chong Ruan, et~al.
\newblock Janus: Decoupling visual encoding for unified multimodal understanding and generation.
\newblock \emph{arXiv preprint arXiv:2410.13848}, 2024.

\bibitem[Wu et~al.(2023)Wu, Fei, Qu, Ji, and Chua]{wu2023next}
Shengqiong Wu, Hao Fei, Leigang Qu, Wei Ji, and Tat-Seng Chua.
\newblock Next-gpt: Any-to-any multimodal llm.
\newblock \emph{arXiv preprint arXiv:2309.05519}, 2023.

\bibitem[Xie et~al.(2024{\natexlab{a}})Xie, Feng, Tian, Lin, Huang, Xia, Gong, Zuo, Yang, Zheng, et~al.]{xie2024learning}
Jinheng Xie, Jiajun Feng, Zhaoxu Tian, Kevin~Qinghong Lin, Yawen Huang, Xi Xia, Nanxu Gong, Xu Zuo, Jiaqi Yang, Yefeng Zheng, et~al.
\newblock Learning long-form video prior via generative pre-training.
\newblock \emph{arXiv preprint arXiv:2404.15909}, 2024{\natexlab{a}}.

\bibitem[Xie et~al.(2024{\natexlab{b}})Xie, Mao, Bai, Zhang, Wang, Lin, Gu, Chen, Yang, and Shou]{xie2024show}
Jinheng Xie, Weijia Mao, Zechen Bai, David~Junhao Zhang, Weihao Wang, Kevin~Qinghong Lin, Yuchao Gu, Zhijie Chen, Zhenheng Yang, and Mike~Zheng Shou.
\newblock Show-o: One single transformer to unify multimodal understanding and generation.
\newblock \emph{arXiv preprint arXiv:2408.12528}, 2024{\natexlab{b}}.

\bibitem[Yang et~al.(2023)Yang, Shen, Zhang, Xu, Zhu, Wu, and Zhou]{yang2023one}
Ceyuan Yang, Yujun Shen, Zhiyi Zhang, Yinghao Xu, Jiapeng Zhu, Zhirong Wu, and Bolei Zhou.
\newblock One-shot generative domain adaptation.
\newblock In \emph{Proceedings of the ieee/cvf international conference on computer vision}, pages 7733--7742, 2023.

\bibitem[Ye et~al.(2024)Ye, Huang, Lu, Yu, Ping, Tao, Kautz, Han, Xu, Molchanov, et~al.]{x-vila}
Hanrong Ye, De-An Huang, Yao Lu, Zhiding Yu, Wei Ping, Andrew Tao, Jan Kautz, Song Han, Dan Xu, Pavlo Molchanov, et~al.
\newblock X-vila: Cross-modality alignment for large language model.
\newblock \emph{arXiv preprint arXiv:2405.19335}, 2024.

\bibitem[You et~al.(2023)You, Guo, Wang, Chang, Baldridge, and Yu]{you2023cobit}
Haoxuan You, Mandy Guo, Zhecan Wang, Kai-Wei Chang, Jason Baldridge, and Jiahui Yu.
\newblock Cobit: A contrastive bi-directional image-text generation model.
\newblock \emph{arXiv preprint arXiv:2303.13455}, 2023.

\bibitem[Yu et~al.(2023{\natexlab{a}})Yu, Lezama, Gundavarapu, Versari, Sohn, Minnen, Cheng, Gupta, Gu, Hauptmann, et~al.]{magvitv2}
Lijun Yu, Jos{\'e} Lezama, Nitesh~B Gundavarapu, Luca Versari, Kihyuk Sohn, David Minnen, Yong Cheng, Agrim Gupta, Xiuye Gu, Alexander~G Hauptmann, et~al.
\newblock Language model beats diffusion--tokenizer is key to visual generation.
\newblock \emph{arXiv preprint arXiv:2310.05737}, 2023{\natexlab{a}}.

\bibitem[Yu et~al.(2023{\natexlab{b}})Yu, Shi, Pasunuru, Muller, Golovneva, Wang, Babu, Tang, Karrer, Sheynin, et~al.]{yu2023scaling}
Lili Yu, Bowen Shi, Ramakanth Pasunuru, Benjamin Muller, Olga Golovneva, Tianlu Wang, Arun Babu, Binh Tang, Brian Karrer, Shelly Sheynin, et~al.
\newblock Scaling autoregressive multi-modal models: Pretraining and instruction tuning.
\newblock \emph{arXiv preprint arXiv:2309.02591}, 2\penalty0 (3), 2023{\natexlab{b}}.

\bibitem[Yu et~al.(2020)Yu, Kumar, Gupta, Levine, Hausman, and Finn]{yu2020gradient}
Tianhe Yu, Saurabh Kumar, Abhishek Gupta, Sergey Levine, Karol Hausman, and Chelsea Finn.
\newblock Gradient surgery for multi-task learning.
\newblock In \emph{Advances in Neural Information Processing Systems}, pages 5824--5836, 2020.

\bibitem[Zhang et~al.(2021)Zhang, Kang, Yang, and Wei]{zhang2021few}
Gengwei Zhang, Guoliang Kang, Yi Yang, and Yunchao Wei.
\newblock Few-shot segmentation via cycle-consistent transformer.
\newblock \emph{Advances in Neural Information Processing Systems}, 34:\penalty0 21984--21996, 2021.

\bibitem[Zhao et~al.(2023)Zhao, Li, Hu, Li, Zou, Shi, and Fan]{zhao2023zero}
Rui Zhao, Wei Li, Zhipeng Hu, Lincheng Li, Zhengxia Zou, Zhenwei Shi, and Changjie Fan.
\newblock Zero-shot text-to-parameter translation for game character auto-creation.
\newblock In \emph{Proceedings of the IEEE/CVF Conference on Computer Vision and Pattern Recognition}, pages 21013--21023, 2023.

\bibitem[Zhao et~al.(2024)Zhao, Gu, Wu, Zhang, Liu, Wu, Keppo, and Shou]{zhao2024motiondirector}
Rui Zhao, Yuchao Gu, Jay~Zhangjie Wu, David~Junhao Zhang, Jia-Wei Liu, Weijia Wu, Jussi Keppo, and Mike~Zheng Shou.
\newblock Motiondirector: Motion customization of text-to-video diffusion models.
\newblock In \emph{European Conference on Computer Vision}, pages 273--290. Springer, 2024.

\bibitem[Zhao et~al.(2025)Zhao, Yuan, Wei, Zhang, Gu, Ran, Wang, Wu, Zhang, Zhang, et~al.]{zhao2025evolvedirector}
Rui Zhao, Hangjie Yuan, Yujie Wei, Shiwei Zhang, Yuchao Gu, Lingmin Ran, Xiang Wang, Jay~Zhangjie Wu, David~Junhao Zhang, Yingya Zhang, et~al.
\newblock Evolvedirector: Approaching advanced text-to-image generation with large vision-language models.
\newblock \emph{Advances in Neural Information Processing Systems}, 37:\penalty0 122104--122129, 2025.

\bibitem[Zhou et~al.(2024)Zhou, Yu, Babu, Tirumala, Yasunaga, Shamis, Kahn, Ma, Zettlemoyer, and Levy]{zhou2024transfusion}
Chunting Zhou, Lili Yu, Arun Babu, Kushal Tirumala, Michihiro Yasunaga, Leonid Shamis, Jacob Kahn, Xuezhe Ma, Luke Zettlemoyer, and Omer Levy.
\newblock Transfusion: Predict the next token and diffuse images with one multi-modal model.
\newblock \emph{arXiv preprint arXiv:2408.11039}, 2024.

\bibitem[Zhu et~al.(2023)Zhu, Chen, Shen, Li, and Elhoseiny]{minigpt4}
Deyao Zhu, Jun Chen, Xiaoqian Shen, Xiang Li, and Mohamed Elhoseiny.
\newblock Minigpt-4: Enhancing vision-language understanding with advanced large language models.
\newblock \emph{CoRR}, abs/2304.10592, 2023.

\bibitem[Zhu et~al.(2017)Zhu, Park, Isola, and Efros]{zhu2017cyclegan}
Jun-Yan Zhu, Taesung Park, Phillip Isola, and Alexei~A. Efros.
\newblock Unpaired image-to-image translation using cycle-consistent adversarial networks.
\newblock In \emph{Proceedings of the IEEE International Conference on Computer Vision (ICCV)}, 2017.

\end{thebibliography}
}


\clearpage
\setcounter{page}{1}
\maketitlesupplementary

\appendix

\section{Model Details}

In this section, we provide an introduction to the base unified generative model, which serves as the foundation for our approach to multimodal understanding and generation. It integrates both auto-regressive and diffusion modeling techniques to achieve joint multimodal understanding and generation~\citep{xie2024show}. 

The base unified generative model is built upon a pre-trained large language model Phi-1.5~\citep{phi1.5}. The architecture of the base model is largely inherited from it with minimal modifications to accommodate multimodal input. Specifically, a QK-Norm operation is added to each attention layer~\citep{qknorm, qknorm2} to enhance training stability. The embedding layer is expanded by adding learnable embeddings for discrete image tokens, enabling the joint encoding of text and image modalities. The final model consists of 24 transformer layers with a total of 1.5 billion parameters. In the proposed DoraCycle, the parameters of the base model are fixed and the LoRA components introduce 4.7 million trainable parameters, accounting for approximately 0.32\% of the total model parameters.

The model tokenizes both text and image data into discrete tokens to create a unified space, maintaining a unified vocabulary.
Text data is tokenized using a pre-trained text tokenizer of Phi-1.5~\citep{phi1.5}. The codebook size of text tokens is 58498.
For images, a quantizer like MAGVIT-v2~\citep{magvitv2} is used. This quantizer maintains a codebook of size $K=8,192$ and encodes images at a resolution of $512\times512$ into $32\times 32$ discrete tokens. 
The model utilizes the unified tokenization strategy ensuring that both modalities can be processed consistently, allowing the model to handle multimodal inputs within a shared framework.
Besides the text and image tokens, the model also involves different special tokens, like \texttt{<sot>}, \texttt{<eot>}, \texttt{<soi>}, and \texttt{<soi>}, which are used to denote the start and end of the text and image tokens. Among them, there are some special tokens that indicate the task to be executed, where the \texttt{<mmu>} indicates the model should do the understanding task and the \texttt{<t2i>} indicates the model should generate image tokens based on the given text.
Additionally, we introduce two new special tokens, \textit{i.e.} \texttt{<soc>}  and \texttt{<eoc>}, to enhance the learning of new concepts.

\section{Differentiable Sampling}
To handle the undifferentiable discrete tokens generation process, we use the Gumbel-Softmax technique to approximate the sampling in a differentiable manner~\citep{jang2016categorical}. The Gumbel-Softmax distribution allows gradients to flow through the sampling process, making it suitable for end-to-end training.

The Gumbel-Softmax operation is defined as follows:

\begin{equation}
y_i = \frac{\exp((\log(\pi_i) + g_i) / \tau)}{\sum_{j=1}^k \exp((\log(\pi_j) + g_j) / \tau)},
\end{equation}
where $\pi_i$ represents the input logits, $g_i$ is sampled from a $Gumbel(0, 1)$ distribution, and $\tau$ is the temperature parameter that controls the softness of the sampling.
The $Gumbel(0, 1)$ distribution is defined as:
\begin{equation}
g_i = -\log(-\log(u_i)),
\end{equation}
where $u_i$ is sampled from a uniform distribution $U(0, 1)$.
In our experiments, we set $\tau = 1.0$ and use the hard Gumbel-Softmax, where the output is discretized to be one-hot, while maintaining the gradient flow for backpropagation.

\section{Training Details}

All experiments are conducted on 8 NVIDIA H100 GPUs, with training taking 3,000 steps. Each 1K steps requires approximately 3.5 hours. We observe that the T Cycle loss typically converges within 1-2K steps. The convergence of I Cycle loss is more challenging to observe directly on the loss curves, while the visual inspection reveals significant convergence.

We employ DeepSpeed Zero with Zero Stage 2, which optimizes memory usage by partitioning model states across devices, along with bf16 mixed precision to efficiently utilize GPU memory and computational resources. The total batch size is set to 32, with each GPU handling a batch size of 4.

For the inference mode to generate pseudo tokens in the middle of each cycle, we set the classifier free guidance to $5$ and generation steps to $30$ for the text-to-image generation and set the Top-K to $1$, temperature to $1.0$, and the max sequence length to $256$ for the image-to-text generation.

\begin{figure*}[!tb]
    \centering
\includegraphics[width=0.935\linewidth]{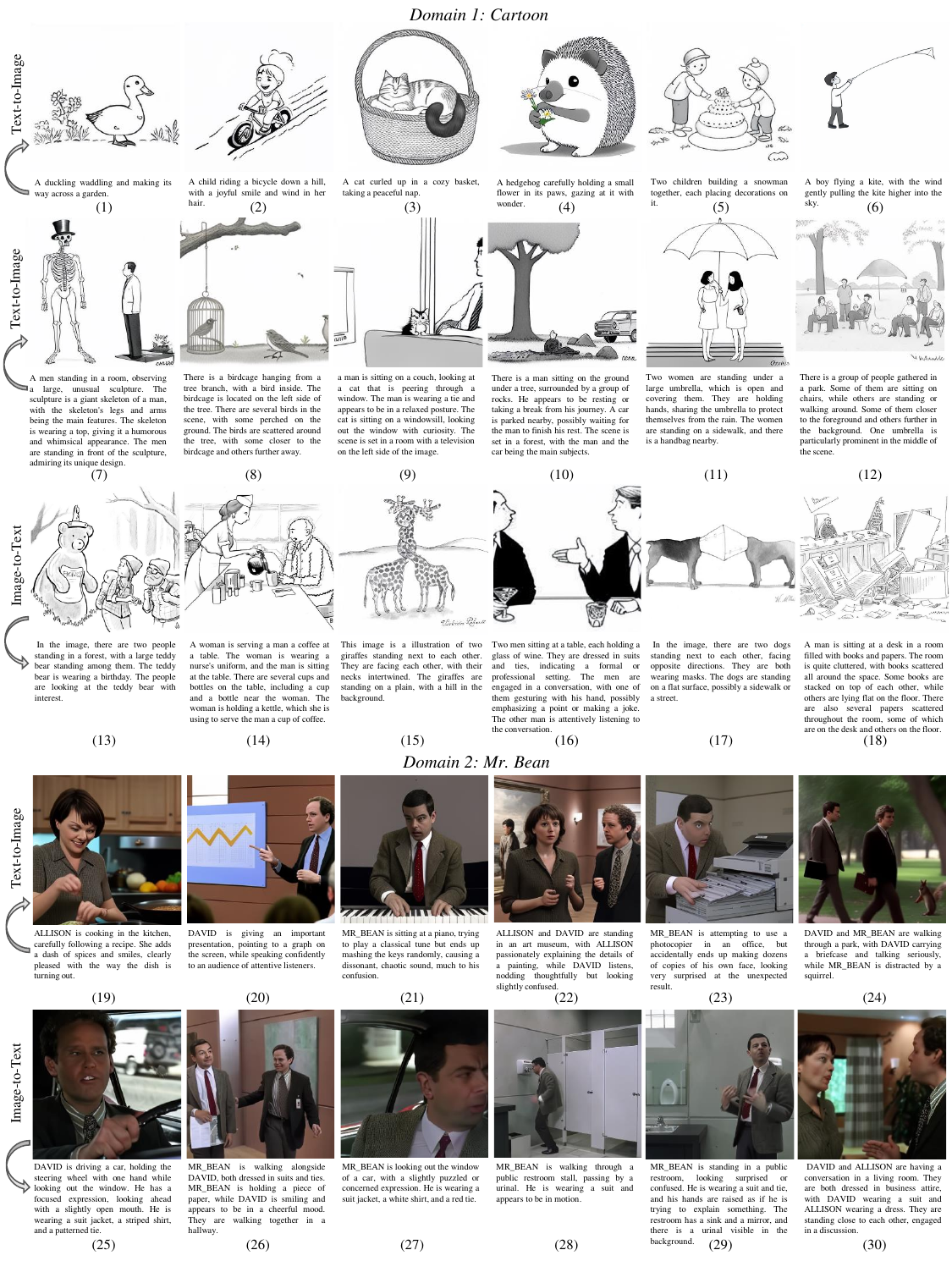}
    \caption{Image-to-text and text-to-image generation by the unified models that adapted for two domains. The special tokens are omitted.}
\label{fig:supp_result_1}
\end{figure*}

\begin{figure*}[!tb]
    \centering
\includegraphics[width=0.95\linewidth]{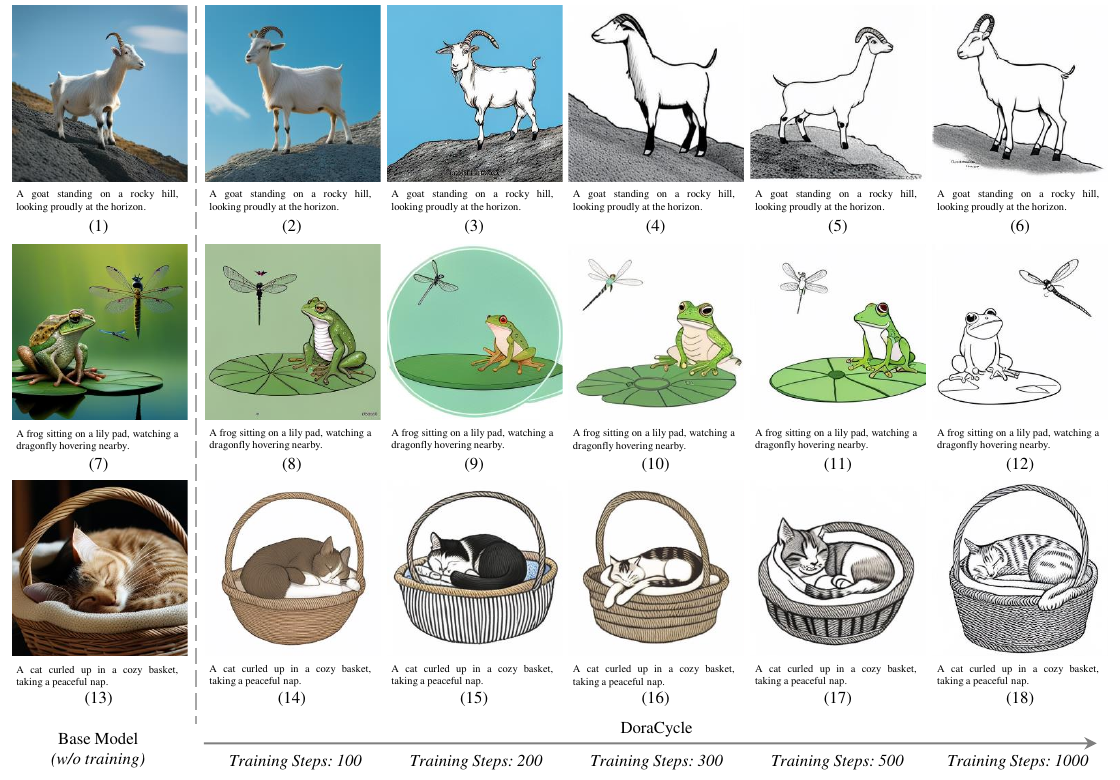}
    \caption{Illustration of the progressive adaptation progress of the unified model to the target domain.}
\label{fig:supp_result_2}
\end{figure*}

\begin{figure*}[!tb]
    \centering
\includegraphics[width=0.95\linewidth]{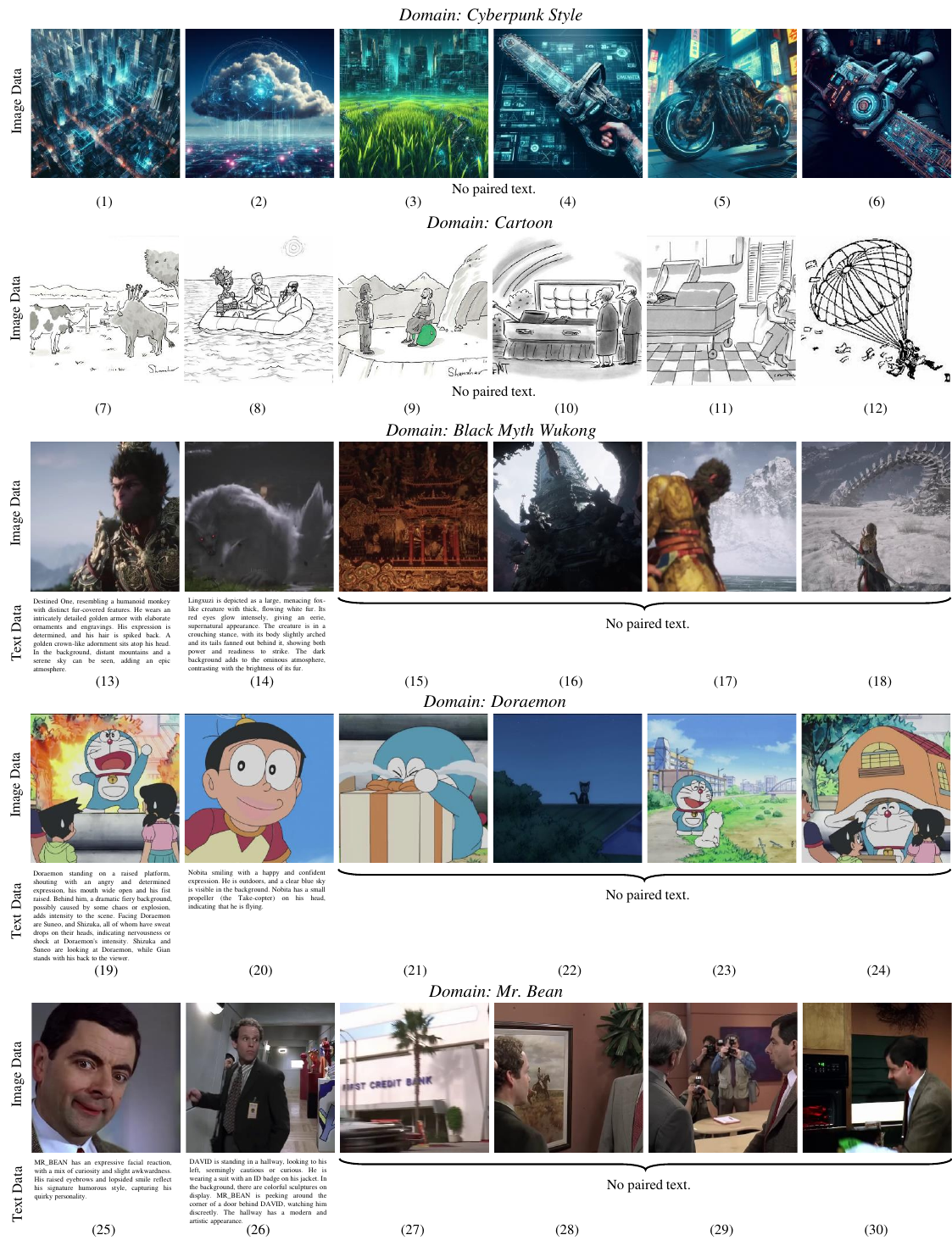}
    \caption{Examples of the training data for DoraCycle to adapt the unified model to different target domains.}
\label{fig:supp_result_3}
\end{figure*}

\section{More Results}
In Fig.~\ref{fig:supp_result_1}, we provide more results generated by the unified model that is adapted by the proposed DoraCycle for two domains, respectively. We omit the special tokens in the figure for brevity.

For domain 1, the cartoon style, which does not need any paired knowledge, we apply DoraCycle on pure unpaired data, collected from the New Yorker Caption Contest Dataset~\citep{hessel2023androids, newyorkernextmldataset, radev-etal-2016-humor, shahaf2015inside}. As shown in  Fig.~\ref{fig:supp_result_1}, the first and second rows present text-to-image generation results. The generated images align well with the characteristics of the target domain, and both short and long text inputs produce well-aligned images. Interestingly, the flowers in Fig.~\ref{fig:supp_result_1} (4) are depicted with bright colors, which can be traced back to the training cartoon data where some objects are highlighted with colors for emphasis, as shown in Fig.~\ref{fig:supp_result_3} (9).
The third row in Figure 1 displays image-to-text generation results, where the model successfully generates accurate text descriptions for images within the domain.

The fourth and fifth rows of Fig.~\ref{fig:supp_result_1} illustrate the adaptation performance of the model to the domain that requires paired knowledge. We collected 2k images from the Mr. Bean movie and annotated each character with 1-3 text descriptions. The adapted model accurately generates images given text prompts, including the character identities, and provides precise descriptions for input images, demonstrating the effective domain adaptation capabilities of the proposed DoraCycle.

In Fig.~\ref{fig:supp_result_2}, we visualize the process of the model gradually adapting to the target domain, \textit{i.e.} the cartoon domain. It can be observed that, given the same text prompt, the images generated by the model become increasingly aligned with the target domain characteristics as training steps increase. Notably, the three sets of examples in Fig.~\ref{fig:supp_result_2} reveal different levels of difficulty in mapping samples to the target domain. For instance, the second set, ``A frog sitting on a lily pad, watching a dragonfly hovering nearby," requires more training to generate an image that fits well within the target domain, while the first set is much easier.

\section{Training Data Examples}
In Fig.~\ref{fig:supp_result_3}, we present some examples of the training data of DoraCycle to adapt the unified generative models to different domains. For domains that do not require paired knowledge, DoraCycle uses purely unpaired data. The rows 1-2 in Fig.~\ref{fig:supp_result_3} show examples of the image data, while the text data uses text from the pre-trained dataset with automatically added domain-specific prompts, such as ``\texttt{<soc>} Cyberpunk Style \texttt{<eoc>}''. For domains that require paired knowledge, we provide a large amount of unpaired data, as shown on the right side of rows 3-5 in Fig.~\ref{fig:supp_result_3}, along with a small amount of paired data (around 1\%), as shown on the left side of rows 3-5 in Fig.~\ref{fig:supp_result_3}.

\section{Limitations and Future Work}
In the current setup, the multimodal cycles only involve text prompts that directly describe visual content, ~\textit{i.e.} captions, limiting the multimodal understanding abilities to caption generation. Extending the framework to include visual question answering is an interesting direction for future work. Directly involving question-answer in the multimodal cycles poses challenges, as the question-answer pair often covers only a specific part of the image, potentially missing many visual details when performing the text-to-image generation process, leading to a large cycle deviation. One potential solution is to leverage the internal language reasoning capabilities of unified models to generate more complete textual descriptions for visual content based on question-answer pair, and then proceed with the multimodal cycles.

\end{document}